\documentclass[10pt,twocolumn,letterpaper]{article}
\usepackage{multirow}
\usepackage{colortbl}
\usepackage{float}
\usepackage[table]{xcolor} % 必须加 [table] 选项
\usepackage[pagenumbers]{cvpr} % To force page numbers, e.g. for an arXiv version

\definecolor{cvprblue}{rgb}{0.21,0.49,0.74}
\usepackage[pagebackref,breaklinks,colorlinks,allcolors=cvprblue]{hyperref}

\def\paperID{} % *** Enter the Paper ID here
\def\confName{CVPR}
\def\confYear{2026}

\title{Expand and Prune: Maximizing Trajectory Diversity for Effective GRPO in Generative Models}

\author{
    Shiran Ge\textsuperscript{1,2,*}
    Chenyi Huang\textsuperscript{1,2,*},
    Yuang Ai\textsuperscript{1,2},
    Qihang Fan\textsuperscript{1,2},
    Huaibo Huang\textsuperscript{1,\dag},
    Ran He\textsuperscript{1,2} 
    \vspace{0.3cm} \\ % 稍微加点垂直间距
    \textsuperscript{1}MAIS \& NLPR, Institute of Automation, CAS \quad 
    \textsuperscript{2}School of Artificial Intelligence, UCAS \\
}

\begin{document}
\maketitle
\begingroup
    \renewcommand\thefootnote{} % 清空符号
    \footnotetext{
    \hspace*{-2\parindent}
    \parbox{\linewidth}{
        *~Equal Contribution. \\
        \dag~Corresponding author: huaibo.huang@cripac.ia.ac.cn \\
        % \noindent
        % % [1.5em] 是盒子的宽度，[l] 代表符号在盒子里左对齐
        % \makebox[1.5em][l]{*}These authors contributed equally. \\
        % \makebox[1.5em][l]{\dag}Corresponding author: huaibo.huang@cripac.ia.ac.cn\\
        % \noindent 
        \textbf{Emails:} % 加个粗体前缀稍微好看点
        \{shiran.ge, rhe\}@nlpr.ia.ac.cn, 
        huangchenyi2026@ia.ac.cn, 
        \{shallowdream555, fanqihang.159\}@gmail.com
        % fanqihang.159@gmail.com
        % rhe@nlpr.ia.ac.cn
    }}
\endgroup
\begin{abstract}
Group Relative Policy Optimization (GRPO) is a powerful technique for aligning generative models, but its effectiveness is bottlenecked by the conflict between large group sizes and prohibitive computational costs. In this work, we investigate the trade-off through empirical studies, yielding two key observations. First, we discover the reward clustering phenomenon in which many trajectories collapse toward the group-mean reward, offering limited optimization value. Second, we design a heuristic strategy named Optimal Variance Filtering (OVF), and verify that a high- variance subset of trajectories, selected by OVF can outperform the larger, unfiltered group. However, this static, post-sampling OVF approach still necessitates critical computational overhead, as it performs unnecessary sampling for trajectories that are ultimately discarded. To resolve this, we propose Pro-GRPO (\textbf{Pro}active GRPO), a novel dynamic framework that integrates latent feature-based trajectory pruning into the sampling process. Through the early termination of reward-clustered trajectories, Pro-GRPO reduces computational overhead. Leveraging its efficiency, Pro-GRPO employs an ``Expand-and-Prune'' strategy. This strategy first expands the size of initial sampling group to maximize trajectory diversity, then it applies multi-step OVF to the latents, avoiding prohibitive computational costs. Extensive experiments on both diffusion-based and flow-based models demonstrate the generality and effectiveness of our Pro-GRPO framework.
\end{abstract}

\section{Introduction}
\label{sec:intro}

Aligning visual generative models with human preferences is crucial for achieving controllable content generation~\cite{cao2023reinforcement,miao2024training,hu2025towards}. Among various alignment techniques, reinforcement learning (RL) methods, such as PPO~\cite{schulman2017proximal}, DPO~\cite{DPO}, and GRPO~\cite{shao2024deepseekmath}, have been widely adopted~\cite{wallace2024diffusion, black2023training,guo2025can,li2025inversion}. Particularly, GRPO is favored for its unique in-group sampling mechanism, which enables robust advantage estimation without the need for a trained value function~\cite{wu2025reprompt,DBLP:journals/corr/abs-2505-17022}. However, the effectiveness of GRPO relies on a large group size to ensure reliable advantage estimation~\cite{liu2025flow}. This requirement imposes memory and computational burdens, forcing a suboptimal trade-off~\cite{xue2025dancegrpounleashinggrpovisual}: while a small group size fits resource constraints, it destabilizes advantage estimation and hampers exploration; conversely, a large group size becomes computationally prohibitive. This dilemma ultimately bottlenecks the model's performance potential.

Delving into this trade-off, we conducted two empirical studies on GRPO’s sampling and obtained two key observations. First, we identify an reward clustering phenomenon: for a single prompt, standard GRPO generates many trajectories whose rewards concentrate near the group mean. This clustering degenerates normalized advantages, limiting the contribution of the corresponding samples to the overall optimization process, while consuming the same computational resources. Meanwhile, naive techniques like random subsampling fail to resolve this phenomenon (Fig.~\ref{fig:reward-clustering}).

Second, building on the observed reward clustering, we hypothesize that such clustering could be detrimental to optimization. To validate this, we propose a simple yet effective strategy named Optimal Variance Filtering (OVF). OVF is a heuristic algorithm that selects trajectories by maximizing reward variance, strategically focusing on both high and low reward extremes. Our experiments show that OVF effectively mitigates reward clustering phenomenon and enhance optimization performance. This validates the \emph{``Less is More''} principle, proving that a smaller, high-variance subset can outperform a larger, unfiltered group.

However, while OVF effectively filters reward-clustered trajectories to enhance optimization by maximizing reward variance, it suffers from the limitation of post-sampling filtering, which still incurs unnecessary computation for reward-clustered trajectories. To address this issue, we introduce Pro-GRPO (Proactive GRPO), a dynamic framework that prunes trajectories based on latent features during sampling. By early-terminating reward-clustered trajectories, the approach significantly improves sampling efficiency. Leveraging this efficiency, Pro-GRPO adopts the ``Expand-and-Prune'' strategy: it first expands the size of initial trajectory sampling group to maximize diversity, and then applies multi-step OVF during the generation process. This approach allows Pro-GRPO to achieve a larger trajectory diversity, and significantly improve performance, without incurring prohibitive computational costs.

In summary, our contributions are:

\begin{enumerate}
    \item We conduct two in-depth empirical studies on GRPO's sampling trajectories, yielding two key observations: 
    (1) GRPO exhibits reward clustering phenomenon. (2) A high-variance subset of trajectories can outperform a larger, unfiltered sampling group.
    \item We introduce the Pro-GRPO framework, which integrates a dynamic latent feature-based pruning framework enabling efficient in-process sampling. Leveraging this efficiency, we adopt an ``Expand-and-Prune'' strategy to maximize trajectory diversity without incurring prohibitive computational costs.
    \item We conduct extensive experiments on both diffusion-based and flow-based generative models, demonstrating the generality and effectiveness of our Pro-GRPO framework.
\end{enumerate}

\section{Related Work}
\label{sec:Relatedwork}

%-------------------------------------------------------------------------
\subsection{Diffusion and Flow Matching Models}
Modern high-fidelity Text-to-Image (T2I) synthesis is primarily driven by diffusion~\cite{podell2023sdxlimprovinglatentdiffusion, betker2023improving, NEURIPS2022_ec795aea, DBLP:conf/cvpr/RuizLJPRA23} and flow matching~\cite{labs2025flux1kontextflowmatching, esser2024scaling, wu2025qwenimagetechnicalreport} models, which share the common objective of learning a continuous path from noise to data distributions~\cite{gao2025diffusion, esser2024scaling}. While diffusion models perform denoising via stochastic differential equation (SDE) solvers or discrete denoising diffusion probabilistic model (DDPM) steps, flow matching directly learns a velocity field, enabling highly efficient deterministic sampling through ordinary differential equation (ODE) steps~\cite{lipman2023flow}. A series of theoretical works~\cite{albergo2023stochastic, gao2025diffusion, albergo2023building, domingo-enrich2025adjoint} have unified the characterization of diffusion and flow matching models under an SDE/ODE framework. This unification provides a theoretical foundation for subsequently integrating online reinforcement learning algorithms, such as GRPO, into the post-training phases of these models~\cite{liu2025flow,li2025mixgrpo, xue2025dancegrpounleashinggrpovisual}.
\subsection{GRPO for T2I Models}
GRPO has gained prominence in aligning T2I models~\cite{guo2025can, wang2025simplear, zhang2025reasongenr1cotautoregressiveimage, gallici2025finetuning, yuan2025argrpo, ma2025stagestablegeneralizablegrpo}, with methods such as FlowGRPO~\cite{liu2025flow} and DanceGRPO~\cite{xue2025dancegrpounleashinggrpovisual} that reformulate the generation process as an equivalent SDE, enabling stochastic policy exploration and facilitating GRPO integration into flow models. Building on these foundations, subsequent research has refined algorithmic details such as optimizing credit assignment~\cite{li2025branchgrpostableefficientgrpo, he2025tempflowgrpotimingmattersgrpo} and improving reward computation~\cite{wang2025prefgrpopairwisepreferencerewardbased}. However, less attention has been paid to reward clustering in sampling groups. Many trajectories concentrate near the group mean, adding little gradient signal. Pro-GRPO tackles this by latent-feature pruning with an expand-and-prune schedule, retaining a small high-variance survivor set that enhance performance without prohibitive cost.
\section{Preliminaries}
\label{sec:preliminaries}

\subsection{Diffusion Models}
A diffusion model~\cite{ho2020denoising} specifies a forward noising process that perturbs a clean sample $\mathbf{x}_1$ toward Gaussian noise. At time $t$ the noisy state is
\begin{equation}
\mathbf{x}_t \;=\; \alpha_t \mathbf{x}_1 + \sigma_t \boldsymbol{\epsilon}, 
\qquad 
\boldsymbol{\epsilon}\sim\mathcal{N}(\mathbf{0},\mathbf{I}),
\label{eq:dm_forward}
\end{equation}
where $(\alpha_t,\sigma_t)$ is the noise schedule.
For sampling, data can be generated from a prior $\mathbf{x}_T \sim \mathcal{N}(\mathbf{0},\mathbf{I})$ by solving the corresponding reverse-time Stochastic Differential Equation (SDE)~\cite{song2019generative,song2020score}:
\begin{equation}
\mathrm{d}\mathbf{x}_t 
= \Big( f(t,\mathbf{x}_t) - \tfrac{1+\eta^{2}}{2}\,\sigma_t^{2}\,\nabla_{\mathbf{x}}\log p_t(\mathbf{x}_t) \Big)\mathrm{d}t 
+ \eta\,\sigma_t\,\mathrm{d}\mathbf{w}_t,
\label{eq:dm_sde}
\end{equation}
where $\mathbf{w}_t$ is a Wiener process, $\nabla_{\mathbf{x}}\log p_t(\mathbf{x}_t)$ is the score, and $\eta$ controls path stochasticity.

\subsection{Rectified Flow}
Rectified Flow (RF)~\cite{liu2022flow} is learned via Flow Matching~\cite{lipman2023flow}, which posits a linear path from $\mathbf{x}_0\!\sim\!\mathcal{N}(\mathbf{0},\mathbf{I})$ to data $\mathbf{x}_1$:
\begin{equation}
\mathbf{x}_t \;=\; (1-t)\mathbf{x}_0 + t\,\mathbf{x}_1,\quad t\in[0,1].
\label{eq:rf_path}
\end{equation}
A velocity field $v_\theta(\mathbf{x}_t,t)$ is trained to approximate $\mathbf{v}=\mathbf{x}_1-\mathbf{x}_0$.
The learned field gives the deterministic probability-flow ODE
$\mathrm{d}\mathbf{x}_t = v_\theta(\mathbf{x}_t,t)\,\mathrm{d}t$.
Following~\cite{albergo2023stochastic,domingo-enrich2025adjoint,liu2025flow}, the deterministic flow can be converted into a reverse SDE that maintains consistent marginal distributions:
\begin{equation}
\mathrm{d}\mathbf{x}_t 
= \Big( v_\theta(\mathbf{x}_t,t) - \tfrac{\sigma_t^{2}}{2}\,\nabla_{\mathbf{x}}\log p_t(\mathbf{x}_t) \Big)\mathrm{d}t 
+ \sigma_t\,\mathrm{d}\mathbf{w}_t.
\label{eq:rf_sde_general}
\end{equation}

\begin{figure*}[t]
  \centering
  \includegraphics[width=\textwidth]{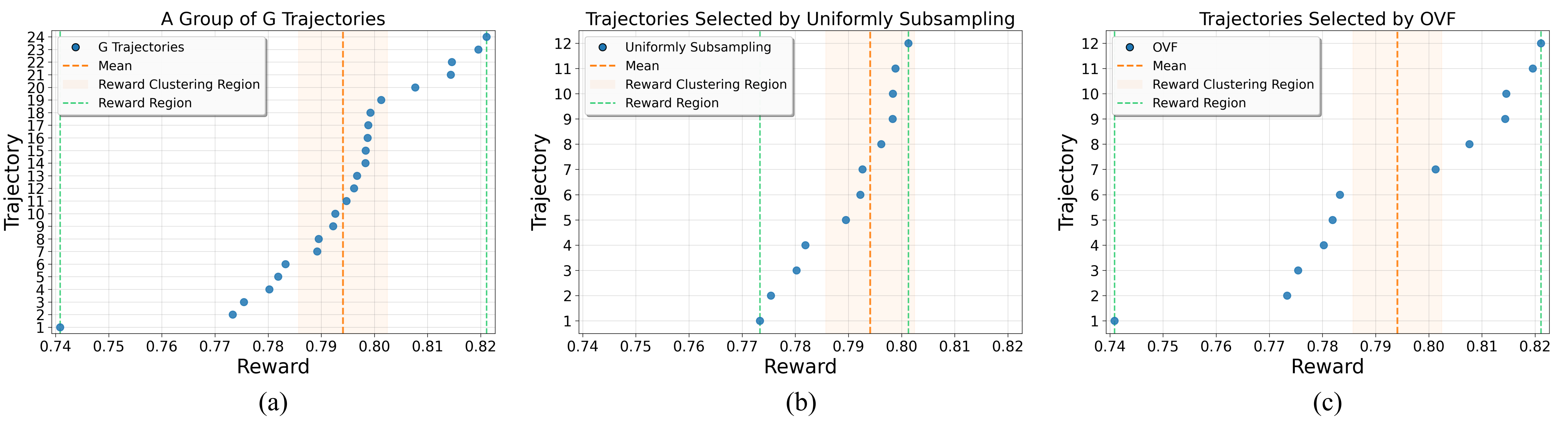}
  
  % --- 核心修改 ---
  \setlength{\abovecaptionskip}{2pt} % 设置图和caption之间的距离，默认通常是10pt左右
  % ---------------
  
  \caption{\textbf{Reward clustering Phenomenon and OVF effects. (a)} A full group ($G=24$) exhibits pronounced reward clustering. \textbf{(b)} Uniform subsampling ($k=12$) preserves the clustering. \textbf{(c)} Our OVF ($k=12$) alleviates reward clustering by selecting from the reward extremes.}
  \label{fig:reward-clustering}
\end{figure*}

\subsection{Group Relative Policy Optimization}
Group Relative Policy Optimization (GRPO)~\cite{shao2024deepseekmath} aligns a pre-trained generator with human preferences using a PPO-style objective~\cite{schulman2017proximal} and a group-relative advantage. 
Given a group of $G$ candidates with rewards $\{R_i\}_{i=1}^{G}$, define
$\mu_G=\tfrac{1}{G}\sum_{j=1}^G R_j$ and 
$\sigma_G=\sqrt{\tfrac{1}{G}\sum_{j=1}^G (R_j-\mu_G)^2}$.
The normalized advantage is
\begin{equation}
A_i \;=\; \frac{R_i-\mu_G}{\sigma_G+\epsilon}, \qquad \epsilon>0,
\label{eq:advantage}
\end{equation}
and the GRPO loss is
\begin{equation}
\begin{aligned}
\mathcal{L}_{\text{GRPO}} &= -\,\mathbb{E}\!\left[ \min\!\big(r\,A,\; \mathrm{clip}(r,\,1-\varepsilon,\,1+\varepsilon)\,A\big) \right] \\
&\quad + \beta\,\mathrm{KL}\!\big(\pi_\theta \,\|\, \pi_{\text{ref}}\big),
\end{aligned}
\end{equation}
where $r$ is the importance ratio between the current and reference policies, 
$\varepsilon$ is the PPO clipping parameter, and $\beta$ weights the KL regularization toward $\pi_{\text{ref}}$.

\begin{figure}[t]
  \centering
  \includegraphics[width=\columnwidth]{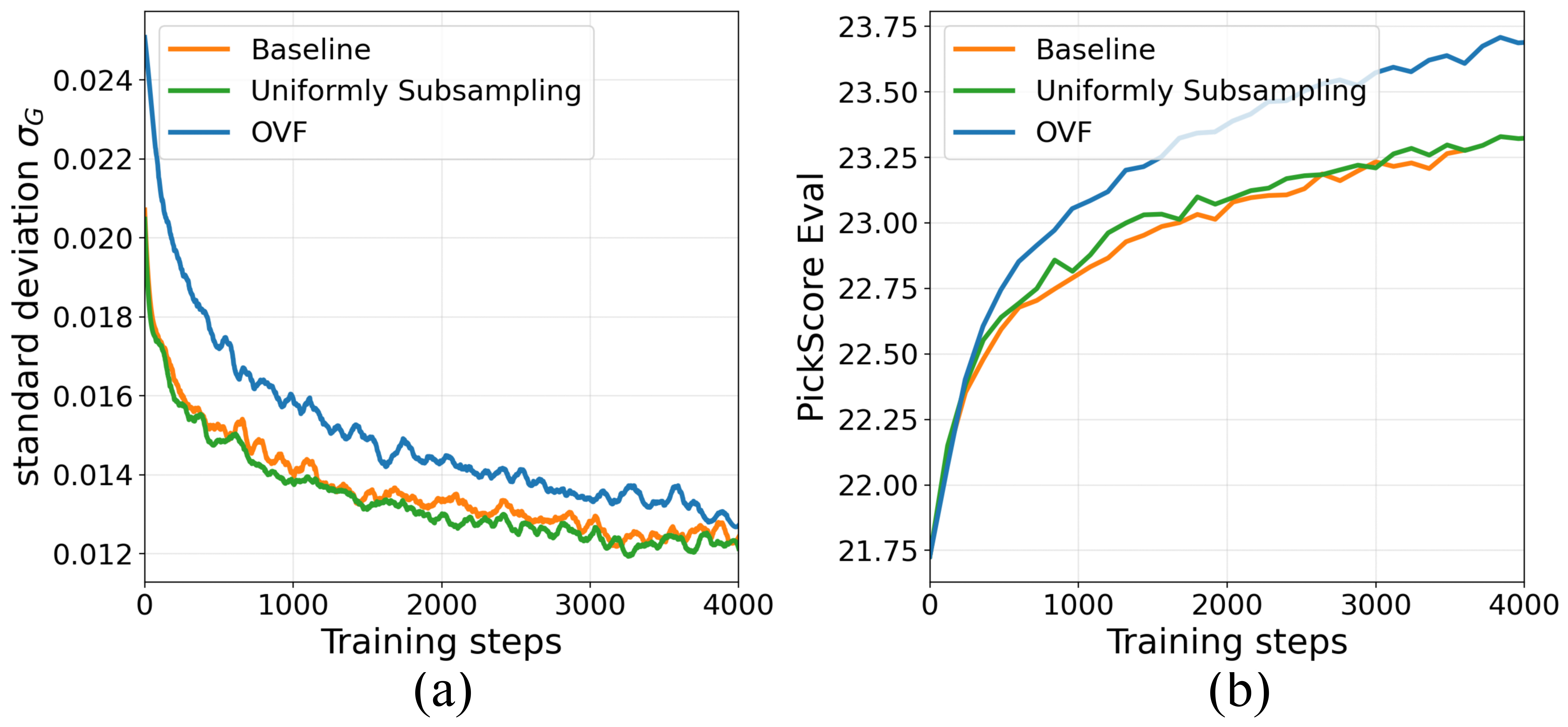} 
  
  \caption{
    \textbf{Visualization of training dynamics on PickScore.}
    We compare the Baseline ($G=24$) against Uniform Subsampling ($k=12$) and our OVF strategy ($k=12$).
    \textbf{(a)} Reward standard deviation ($\sigma_G$).
    \textbf{(b)} PickScore Eval.
  }
  \setlength{\abovecaptionskip}{2pt}
  \label{fig:ovf-training-curves} 
\end{figure}

\section{Trajectory Analyses of GRPO}
\label{sec:empirical}

\subsection{Reward Clustering Phenomenon}
\label{sec:clustering}
The effectiveness of GRPO is closely related to the reward distribution within its sampling group~\cite{liu2025flow}. To resolve the conflict between effectiveness and computational cost, we first examine this distribution empirically.

For a single prompt, we draw a group of $G$ trajectories under the reference policy and compute rewards $\{R_i\}_{i=1}^G$ with group mean $\mu_G$ and standard deviation $\sigma_G$. As illustrated in Fig.~\ref{fig:reward-clustering} (a), we observe a reward clustering phenomenon: a large fraction of rewards concentrate near $\mu_G$, leading to a small $\sigma_G^2$. We formalize the clustered region by
\begin{equation}
\mathcal{C}_\delta \,=\, \big\{\, i \;:\; |R_i-\mu_G| \le \delta\,\sigma_G \,\big\},
\label{eq:near-mean}
\end{equation}
where $\delta>0$ controls the neighborhood width.

This clustering induces an attenuation of the advantage signal. For any $i\in\mathcal{C}_\delta$, substituting the advantage definition (Eq.~\eqref{eq:advantage}) into \eqref{eq:near-mean} yields a mathematical bound
\begin{equation}
|A_i|
= \frac{|R_i-\mu_G|}{\sigma_G+\epsilon}
\le \frac{\delta\,\sigma_G}{\sigma_G+\epsilon}
\le \delta,
\label{eq:adv-bound}
\end{equation}
showing that as the within-group variance $\sigma_G^2$ shrinks, the normalized advantage of clustered samples vanishes.

The per-trajectory gradient contribution scales with the advantage,
\begin{equation}
g_i \;\propto\; A_i \,\nabla_\theta \log \pi_\theta(\tau_i).
\end{equation}
and the mini-batch gradient satisfies $g = \tfrac{1}{G}\sum_{i=1}^G g_i$. Hence, when $A_i\!\approx\!0$ for many $i\in\mathcal{C}_\delta$, those trajectories contribute negligibly to $g$ despite consuming the same compute.

Finally, a naive solution like uniformly subsampling $k<G$ candidates fails to resolve this. As shown in Fig.~\ref{fig:reward-clustering}(b), random subsampling does not increase the variance in expectation; it merely preserves the near-mean concentration. This leaves both the magnitude of $A$ and the effective gradient remain limited.

\begin{figure*}[!htbp] % 或 [!t]
  \centering
  \includegraphics[width=\textwidth]{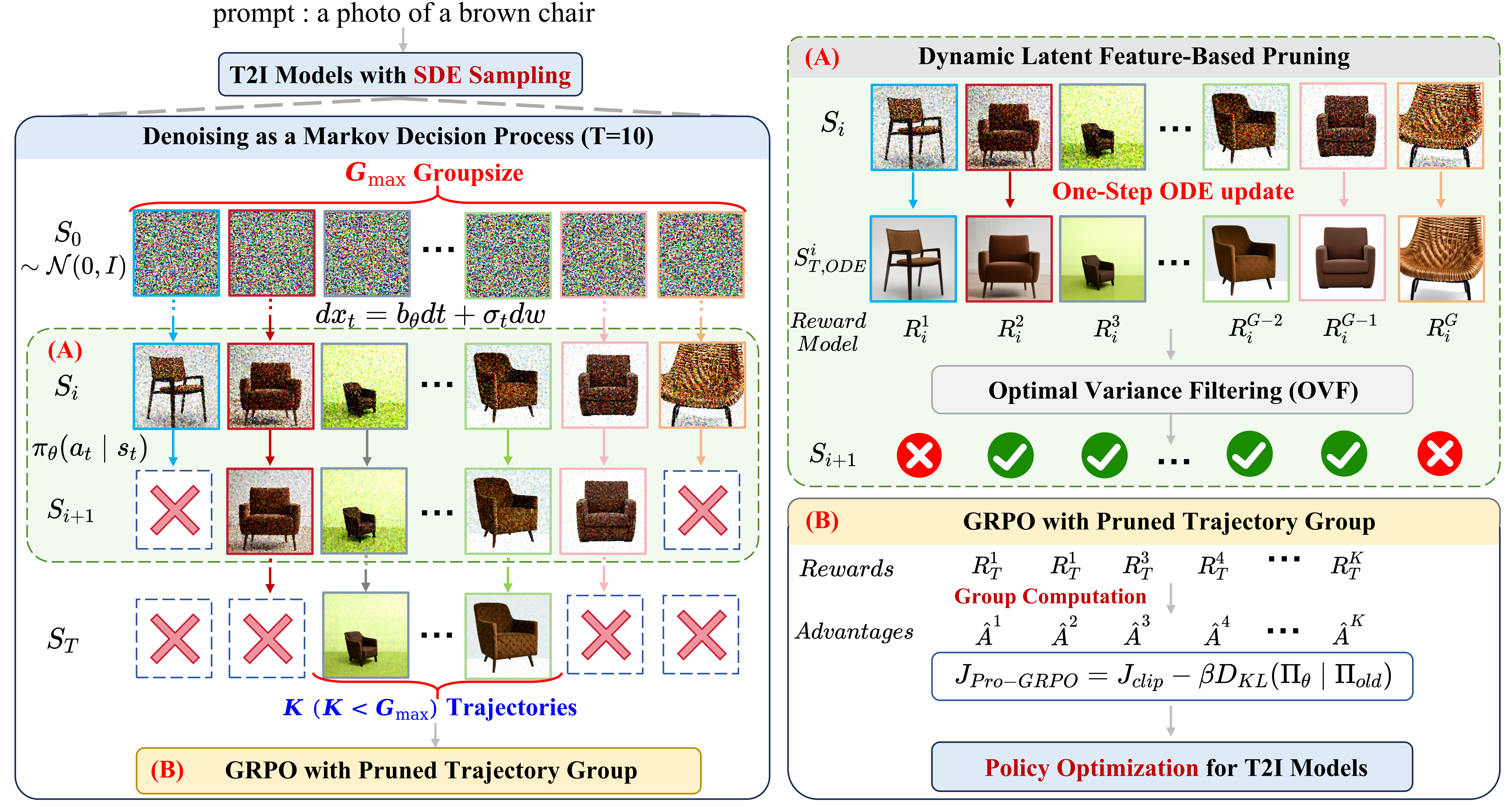} 
  \caption{
    \textbf{Overview of the Pro-GRPO.}
Pro-GRPO runs a dynamic expand-and-prune schedule inside the \(T\)-step denoising.
We begin with an expanded group \(G_{\max}\) to maximize exploration. 
At intermediate checkpoints \(S_i\), \textbf{(A) Dynamic latent pruning} deterministically projects each active latent to \(T\) via a single-step ODE update to obtain a proxy sample and reward \(\hat{R}_i\); OVF then keeps a high-variance subset and early-terminates the rest (red crosses), progressively narrowing the group to \(K\) survivors at \(S_T\).
\textbf{(B) GRPO on the pruned group} computes group-normalized advantages over the \(K\) survivors and updates the policy, achieve high performance with low computational cost.
  }
  \setlength{\abovecaptionskip}{2pt}
  \label{fig:progrpo}
\end{figure*}

\subsection{Optimal Variance Filtering}
\label{sec:OVF}

Building on the observed reward clustering in Section~\ref{sec:clustering}, we hypothesize that such clustering is detrimental to optimization. To validate this, we propose Optimal Variance Filtering (OVF), a simple yet effective strategy designed to amplify the advantage signal by maximizing the within-group reward variance. 
Given rewards $\{R_g\}_{g=1}^{G}$ and a target size $k<G$, OVF selects the subset
\begin{equation}
\mathcal{K}^\star
=
\arg\max_{\substack{\mathcal{K}\subseteq\{1,\dots,G\}\\ |\mathcal{K}|=k}}
\sigma^2(\mathcal{K}).
\label{eq:ovf}
\end{equation}

By construction, Eq.~\eqref{eq:ovf} prefers subsets that span the reward extremes, thus mitigating the reward clustering phenomenon mentioned in \cref{sec:clustering}.

The effect of OVF is twofold. First, as visualized in Fig.~\ref{fig:reward-clustering} (c), OVF-selected rewards are de-clustered; the distribution spreads toward both tails, confirming the mechanism is effective at mitigating clustering. Second, we evaluate the long-term training impact in Fig.~\ref{fig:ovf-training-curves}. Fig.~\ref{fig:ovf-training-curves} (a) shows that OVF (blue) consistently maintains a higher reward variance ($\sigma_G$) throughout training, while the variance of the Baseline (orange) and Uniform Subsampling (green) groups remains comparatively lower. Fig.~\ref{fig:ovf-training-curves} (b) demonstrates that training on this OVF-filtered subset ($k < G$) does not just avoid degradation, but outperforms the baseline trained on the full, unfiltered group ($G$).

This result confirms our ``Less is More'' hypothesis: a smaller, high-variance subset is not just sufficient, but can outperform a larger, unfiltered group.

\section{Pro-GRPO}
\label{sec:pro-grpo}

While OVF demonstrates that maximizing within-group reward variance improves learning, it is inherently a post-sampling filter: reward-clustered trajectories must be fully generated before selection. We therefore introduce Pro-GRPO, a dynamic framework that prunes during sampling. It uses latent features to early-terminate reward-clustered trajectories and employs an “Expand-and-Prune” strategy. The overall pipeline is shown in Fig.~\ref{fig:progrpo}.

\subsection{Dynamic Pruning Framework}
\label{sec:pruning}

\noindent\textbf{Initialization and multi-step denoising.}
Given a prompt $c$, we draw $G$ initial latents $\{\mathbf{x}^{(g)}_{0}\}_{g=1}^{G}$ from $\mathcal{N}(\mathbf{0},\mathbf{I})$ and run reverse-time denoising for 
$T$ steps. As shown in Section \cref{sec:preliminaries}, for both diffusion and rectified flow (RF) models, one reverse-time sampling step admits the unified SDE
\begin{equation}
\mathrm{d}\mathbf{x}_t
= b_{\theta}(\mathbf{x}_t,t)\,\mathrm{d}t
+ \sigma_t\,\mathrm{d}\mathbf{w}_t ,
\label{eq:unified_sde_method}
\end{equation}
where $\mathbf{w}_t$ is a $d$-dimensional Wiener process and $\sigma_t\!\ge\!0$ controls stochasticity. The drift specializes to each backbone as
\begin{equation}
b_{\theta}(\mathbf{x},t)=
\begin{cases}
f(t,\mathbf{x}) - \tfrac{1+\eta^2}{2}\,\sigma_t^2\,\nabla_{\mathbf{x}}\log p_t(\mathbf{x}), & \text{Diffusion},\\[4pt]
v_{\theta}(\mathbf{x},t) - \tfrac{1}{2}\,\sigma_t^2\,\nabla_{\mathbf{x}}\log p_t(\mathbf{x}), & \text{RF}.
\end{cases}
\label{eq:backbone_drifts_method}
\end{equation}
With Euler–Maruyama discretization and step size $\Delta t$, the $g$-th trajectory updates as follows, where $\boldsymbol{\xi}^{(g)}_{t}\sim\mathcal{N}(\mathbf{0},\mathbf{I})$.
\begin{equation}
\mathbf{x}^{(g)}_{t+\Delta t}
= \mathbf{x}^{(g)}_{t} + b_{\theta}\!\left(\mathbf{x}^{(g)}_{t},t\right)\Delta t
+ \sigma_t\sqrt{\Delta t}\,\boldsymbol{\xi}^{(g)}_{t}.
\label{eq:euler_method}
\end{equation}

\noindent\textbf{Dynamic Latent Feature-Based Pruning.}
At pre-specified checkpoints $t_i$, we cheaply predict the terminal outcome for each active trajectory by computing a single-step deterministic ODE projection. We use the probability-flow ODE drift
\begin{equation}
b^{\mathrm{ODE}}_{\theta}(\mathbf{x},t)=
\begin{cases}
f(t,\mathbf{x}) - \tfrac{1}{2}\sigma_t^2 \nabla_{\mathbf{x}}\log p_t(\mathbf{x}), & \text{Diffusion},\\[4pt]
v_{\theta}(\mathbf{x},t), & \text{RF},
\end{cases}
\label{eq:ode_drift_method}
\end{equation}
and approximate the ODE terminal preview from $\mathbf{x}^{(g)}_{t_i}$ by one Euler step:
\begin{equation}
\mathbf{x}^{(g)}_{T,\mathrm{ODE}}
\;\approx\;
\mathbf{x}^{(g)}_{t_i} + (T-t_i)\, b^{\mathrm{ODE}}_{\theta}\!\big(\mathbf{x}^{(g)}_{t_i},\,t_i\big).
\label{eq:ode_lookahead_method}
\end{equation}
Next, we pass this preview latent $\mathbf{x}_{T,\mathrm{ODE}}^{(g)}$ through the VAE decoder and the reward model $R(\cdot, c)$ to obtain a proxy reward
\begin{equation}
R^{(g)}_{i} \;=\; R\!\left(\mathbf{x}^{(g)}_{T,\mathrm{ODE}},\, c\right).
\label{eq:proxy_reward_method}
\end{equation}

Let $K_i$ denote the number of active trajectories at checkpoint $t_i$ (with $K_1=G$).
Optimal Variance Filtering (OVF) selects a subset
$\mathcal{K}_{i+1}\subseteq\{1,\dots,K_i\}$ of size $K_{i+1}<K_i$
that maximizes the within-set reward variance:
\begin{equation}
\mathcal{K}_{i+1}
=
\arg\max_{\substack{\mathcal{K}\subseteq\{1,\dots,K_i\}\\ |\mathcal{K}|=K_{i+1}}}
\frac{1}{K_{i+1}}\sum_{g\in\mathcal{K}}\!\left(R^{(g)}_{i}-\mu_{\mathcal{K}}\right)^{2}.
\end{equation}

Only trajectories indexed by $\mathcal{K}_{i+1}$ continue denoising after $t_i$;
the rest are early-terminated and incur no further SDE steps.
Repeating this procedure across all checkpoints yields a decreasing sequence of survivor counts
$K_1=G > K_2 > \cdots > K_{I+1}=K$.

\noindent\textbf{GRPO with the pruned trajectory set.}
At terminal time $T$, we compute final rewards for the survivors $g\in\mathcal{K}_{I+1}$ and calculate the group-normalized advantages:
\begin{equation}
\widehat{A}^{(g)} = \frac{R(\mathbf{x}^{(g)}_{T},c) - \mu_K}{\sigma_K + \epsilon},
\label{eq:advK}
\end{equation}
where $\mu_K$ and $\sigma_K$ represent the mean and standard deviation of the rewards $\{R(\mathbf{x}^{(h)}_{T},c)\}_{h\in\mathcal{K}_{I+1}}$ within the pruned set.

Let $\mathbf{a}^{(g)}_t=\mathbf{x}^{(g)}_{t+\Delta t}-\mathbf{x}^{(g)}_{t}$ and $\mathbf{s}^{(g)}_t=(\mathbf{x}^{(g)}_{t},t)$. The per-step importance ratio is
\begin{equation}
r^{(g)}_t(\theta)=
\frac{\pi_{\theta}\!\big(\mathbf{a}^{(g)}_t\mid \mathbf{s}^{(g)}_t\big)}
{\pi_{\mathrm{old}}\!\big(\mathbf{a}^{(g)}_t\mid \mathbf{s}^{(g)}_t\big)} ,
\label{eq:ratio_method}
\end{equation}
where $\pi_{\theta}$ is the Gaussian policy induced by Eq.~\eqref{eq:euler_method}. We then maximize the Pro-GRPO objective restricted to the surviving trajectories:
\begin{equation}
\begin{aligned}
&\mathcal{J}_{\mathrm{Pro\text{-}GRPO}}(\theta)
= \mathbb{E}_{c}\!\Big[
\frac{1}{K}\sum_{g\in\mathcal{K}_{I+1}}
\frac{1}{T}\sum_{t=0}^{T-1}
\min\!\big( r^{(g)}_t(\theta)\,\widehat{A}^{(g)}, \\
&\operatorname{clip}\big(r^{(g)}_t(\theta),\,1-\varepsilon,\,1+\varepsilon\big)\,\widehat{A}^{(g)}\big)
\Big]
\;-\; \beta\, D_{\mathrm{KL}}\!\big(\pi_{\theta}\,\|\,\pi_{\mathrm{ref}}\big).
\end{aligned}
\label{eq:pro_grpo_method}
\end{equation}

In this way, Pro-GRPO operates at the optimization cost of a small group of size $K$, dramatically improving sampling efficiency while enhancing performance.

\subsection{Expand and Prune}
\label{sec:expand}

Leveraging the efficiency of dynamic pruning, Pro-GRPO implements an "Expand-and-Prune" strategy that broadens exploration without incurring prohibitive cost. Instead of starting from a budget-constrained group, Pro-GRPO temporarily expands the initial trajectory pool to a larger size \(G_{\max}\! > \!K\) to increase coverage of the reward landscape. During generation, pruning is applied in-process: at each checkpoint \(t_i\) (\(0<t_1<\cdots<t_I<T\)), given \(K_i\) active trajectories, OVF (Eq.~\eqref{eq:ovf}) selects a high-variance subset \(\mathcal{K}_{i+1}\) of size \(K_{i+1}<K_i\) to continue, while the rest are early-terminated. Repeating this selection yields a monotone funnel of survivor counts \(K_1=G_{\max} > K_2 > \cdots > K_{I+1}=K\). In effect, Pro-GRPO expands the exploration benefits of a large initial group yet keeps the effective integration and optimization cost close to that of the final survivors, thereby delivering superior learning signals under a fixed compute budget.

\section{Experiments}
\label{sec:Experiments}

\subsection{Experimental Setup}
\label{sec:exp-setup}

\noindent\textbf{Baselines.} 
To demonstrate the applicability of Pro-GRPO, we conduct experiments across two generative paradigms: the diffusion-based Stable Diffusion v1.4 (SD-v1.4)~\cite{rombach2022high} and the flow-based Stable Diffusion 3.5 (SD3.5-M)~\cite{esser2024scaling}. We compare our framework against representative RL fine-tuning methods, specifically Flow-GRPO~\cite{liu2025flow} and DanceGRPO~\cite{xue2025dancegrpounleashinggrpovisual}. For a fair comparison, all methods are optimized using standard reward objectives, including HPS-v2.1~\cite{wu2023better}, CLIP Score~\cite{radford2021learning}, and PickScore~\cite{kirstain2023pick}, ensuring that performance gains are attributable to the training strategy rather than the reward signal.

\noindent\textbf{Evaluation Metrics.}
We evaluate performance on two standard benchmarks: DrawBench~\cite{saharia2022photorealistic} and HPSv2~\cite{wu2023human}. DrawBench is a benchmark with diverse prompts to evaluate model capabilities, while HPSv2 is designed to evaluate how well generated images align with human preferences.
To provide a comprehensive assessment beyond task-specific accuracy, we evaluate the generated images using a suite of metrics, including ImageReward~\cite{xu2023imagereward}, HPS-v2.1~\cite{wu2023better}, PickScore~\cite{kirstain2023pick}, GenEval~\cite{ghosh2023geneval}, and Aesthetic Score~\cite{schuhmann2022laion}. The results of these evaluations, presented in Section~\ref{sec:main-results}, demonstrate the effectiveness of our Pro-GRPO pipeline in improving both alignment and performance across multiple metrics.

\begin{figure*}[t]
  \centering
  % --- 请替换为您的图片路径 ---
  \includegraphics[width=\textwidth]{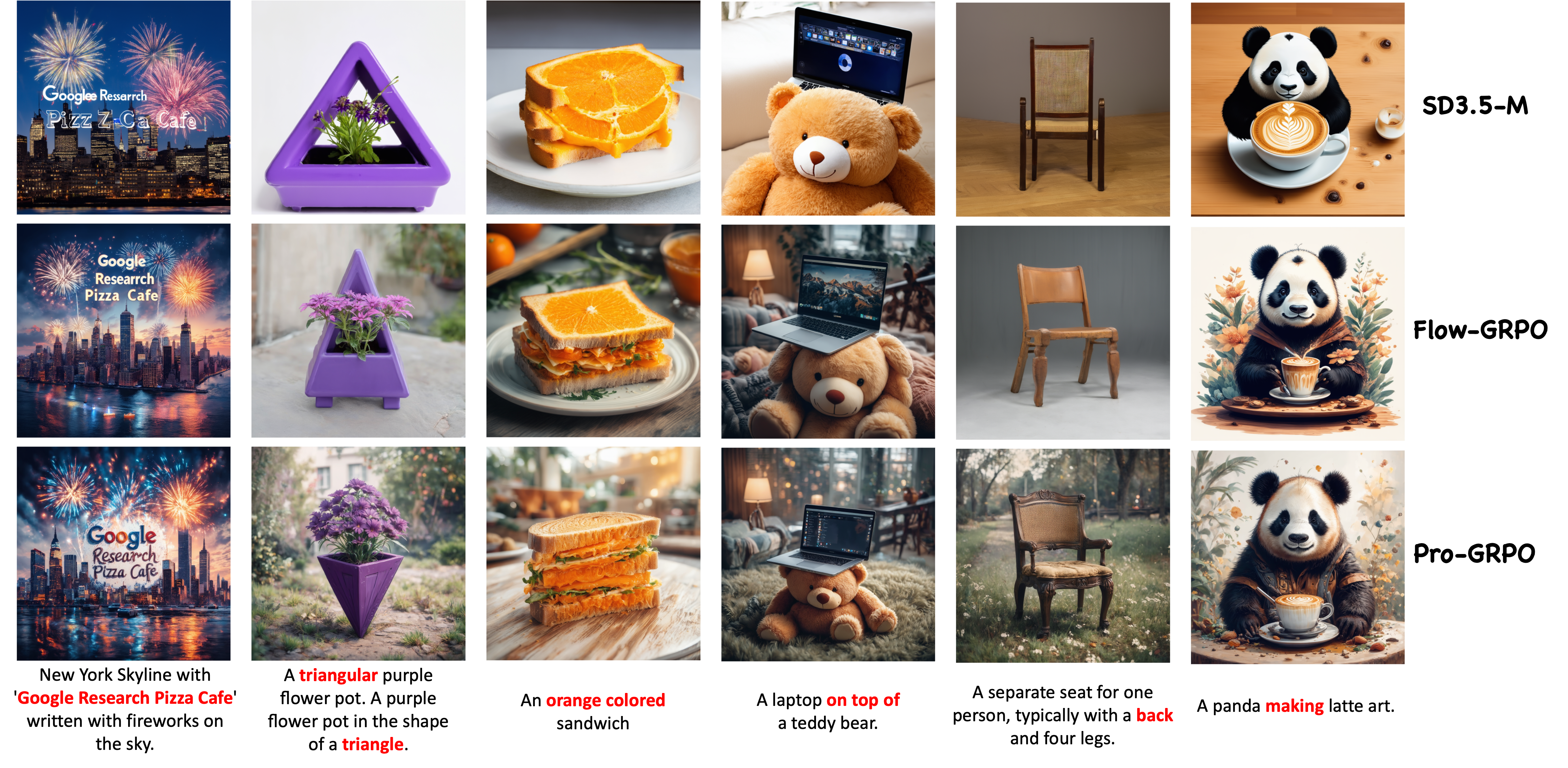}
  
  \caption{
  Qualitative comparison between SD3.5-M, Flow-GRPO and Pro-GRPO with Pickscore as reward on DrawBench prompts.
  }
  \setlength{\abovecaptionskip}{2pt}
  \label{fig:qualitative_results}
\end{figure*}

\begin{table*}[t]
    \centering
    \caption{\textbf{Quantitative comparison on flow-based text-to-image generation (SD3.5-Medium).} We compare Pro-GRPO against the base model and the Flow-GRPO baseline. Note that \textbf{Pro-GRPO-Flash} achieves significant speedup while surpassing the baseline, and \textbf{Pro-GRPO (Standard)} achieves the best overall performance with moderate acceleration. \textbf{Bold} indicates the best result.}
    \label{tab:flow_results}
    \resizebox{0.98\textwidth}{!}{
        \begin{tabular}{cl c c cccc} % 增加了一列 c
            \toprule
            \multirow{2}{*}{\textbf{Dataset}} & \multicolumn{1}{c}{\multirow{2}{*}{\textbf{Model}}} & \textbf{Efficiency} & \textbf{In-Domain} & \multicolumn{4}{c}{\textbf{Out-of-Domain}} \\
            % 增加 cmidrule
            \cmidrule(lr){3-3} \cmidrule(lr){4-4} \cmidrule(lr){5-8}
             & & Speedup ($\uparrow$) & PickScore ($\uparrow$) & Aes. ($\uparrow$) & IR ($\uparrow$) & Pick. ($\uparrow$) & HPSv2.1 ($\uparrow$) \\
            \midrule
            % --- DrawBench Block ---
            \multirow{4}{*}{\textbf{DrawBench}} 
             & SD 3.5-M (Base) & - & - & 5.408 & 0.852 & 22.425 & 0.280 \\
             & Flow-GRPO & $1.00\times$ & 23.322 & 5.912 & 1.298 & 23.599 & 0.316 \\
             & Pro-GRPO-Flash & \textbf{1.41$\times$} & 23.722 & 6.030 & 1.381 & 23.868 & 0.319 \\
             % Pro-GRPO (Ours)
             \rowcolor{gray!10} \cellcolor{white} & \textbf{Pro-GRPO (Ours)} & 1.26$\times$ & \textbf{24.008} & \textbf{6.046} & \textbf{1.397} & \textbf{24.108} & \textbf{0.322} \\
            \midrule
            % --- HPSv2 Block ---
            \multirow{4}{*}{\textbf{HPSv2}} 
             & SD 3.5-M (Base) & - & - & 5.868 & 1.174 & 22.608 & 0.308 \\
             & Flow-GRPO & $1.00\times$ & 23.322 & 6.232 & 1.463 & 23.899 & 0.338 \\
             & Pro-GRPO-Flash & \textbf{1.41$\times$} & 23.722 & 6.303 & 1.495 & 24.251 & 0.340 \\
             % Pro-GRPO (Ours)
             \rowcolor{gray!10} \cellcolor{white} & \textbf{Pro-GRPO (Ours)} & 1.26$\times$ & \textbf{24.008} & \textbf{6.375} & \textbf{1.509} & \textbf{24.597} & \textbf{0.344} \\
            \bottomrule
        \end{tabular}
    }
\end{table*}

\noindent\textbf{Experimental Settings.}
We utilize 24 A100 GPUs for flow-based models and 8 A100 GPUs for diffusion-based models. Our configurations are meticulously aligned with the baselines. 
Against Flow-GRPO ($G=24, T=10$), we evaluate Pro-GRPO (Standard), which expands $G_{\text{max}}=48$ and prunes at steps $t=\{5, 7\}$ (path: $48 \to 24 \to 12$), alongside the efficiency variant Pro-GRPO (Flash) ($G_{\text{max}}=24$, path: $24 \to 16 \to 12$). 
Against DanceGRPO ($G=16, T=50$), we adapt our schedule to start with $G_{\text{max}}=48$ and prune at $t=\{30, 40\}$ (path: $48 \to 32 \to 8$). 
Full configurations are detailed in the Appendix.

\subsection{Main Results}
\label{sec:main-results}

\begin{figure*}[t]
  \centering
  % --- 请替换为您的图片路径 ---
  \includegraphics[width=\textwidth]{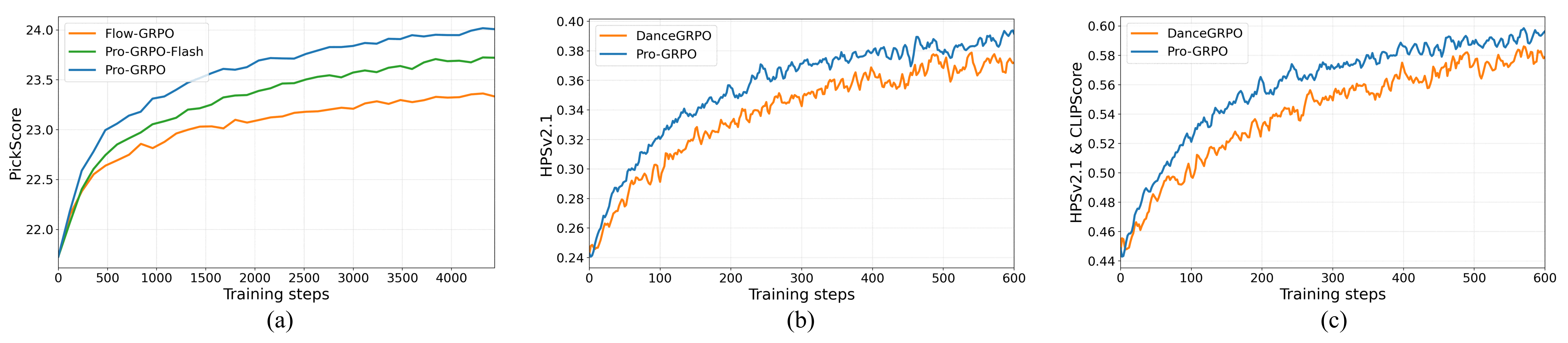}
\caption{\textbf{Training dynamics.}
Reward trajectories during optimization.
\textbf{(a) Flow-based (SD3.5, PickScore):} Pro-GRPO (blue) and Pro-GRPO-Flash (green) converge faster and reach higher plateaus than Flow-GRPO (orange).
\textbf{(b) Diffusion-based (SD-v1.4, HPSv2.1):} Pro-GRPO consistently outperforms DanceGRPO throughout training.
\textbf{(c) Diffusion-based (SD-v1.4, HPSv2.1 \& CLIP):} Pro-GRPO maintains a stable margin, indicating stronger multi-objective optimization.}
  \setlength{\abovecaptionskip}{2pt}
  \label{fig:training_curves}
\end{figure*}

\begin{table*}[t]
    \centering
    \caption{\textbf{Quantitative comparison on diffusion-based T2I generation (SD-v1.4).} We compare Pro-GRPO against DanceGRPO under single-objective (HPSv2.1) and multi-objective (HPSv2.1 \& CLIP) rewards. ``In-Domain'' refers to the metric used during optimization (HPSv2.1 \& CLIP), while ``Out-of-Domain'' metrics assess generalization. \textbf{Bold} indicates the best performance.}
    \label{tab:diffusion_results}
    \resizebox{0.95\textwidth}{!}{
        \begin{tabular}{clcccccc}
            \toprule
            \multirow{2}{*}{\textbf{Dataset}} & \multicolumn{1}{c}{\multirow{2}{*}{\textbf{Model}}} & \multicolumn{2}{c}{\textbf{In-Domain}} & \multicolumn{4}{c}{\textbf{Out-of-Domain}} \\
            \cmidrule(lr){3-4} \cmidrule(lr){5-8}
             & & HPSv2.1 ($\uparrow$) & CLIP ($\uparrow$) & Aes. ($\uparrow$) & IR ($\uparrow$) & Pick. ($\uparrow$) & HPSv2.1 ($\uparrow$) \\
            \midrule
            % --- DrawBench Block ---
             & SD v1.4 (Base) & - & - & 5.207 & -0.029 & 21.084 & 0.2445 \\
             & DanceGRPO (HPS) & 0.369 & - & 5.766 & 0.884 & \textbf{21.903} & 0.3250 \\
             % Pro-GRPO (HPS): 灰色背景, HPS分数加粗 (0.391 > 0.369)
             \rowcolor{gray!10} \cellcolor{white} & \textbf{Pro-GRPO (HPS)} & \textbf{0.391} & - & \textbf{5.894} & \textbf{0.933} & 21.896 & \textbf{0.3390} \\
             % Dance (Combined): 0.575 居中
             & DanceGRPO (HPS+CLIP) & \multicolumn{2}{c}{0.575} & 5.821 & 0.835 & 21.875 & 0.3230 \\
             % Pro-GRPO (Combined): 灰色背景, 0.596 居中加粗, DrawBench标签定义
             \rowcolor{gray!10} \cellcolor{white} \multirow{-5}{*}{\textbf{DrawBench}} & \textbf{Pro-GRPO (HPS+CLIP)} & \multicolumn{2}{c}{\textbf{0.596}} & \textbf{5.897} & \textbf{0.974} & \textbf{22.073} & \textbf{0.3390} \\
            \midrule
            % --- HPSv2 Block ---
             & SD v1.4 (Base) & - & - & 5.389 & 0.015 & 20.688 & 0.2390 \\
             & DanceGRPO (HPS) & 0.369 & - & 6.229 & 1.079 & 22.127 & 0.3660 \\
             % Pro-GRPO (HPS)
             \rowcolor{gray!10} \cellcolor{white} & \textbf{Pro-GRPO (HPS)} & \textbf{0.391} & - & \textbf{6.253} & \textbf{1.114} & \textbf{22.202} & \textbf{0.3830} \\
             & DanceGRPO (HPS+CLIP) & \multicolumn{2}{c}{0.575} & 6.212 & 1.057 & 22.102 & 0.3670 \\
             % Pro-GRPO (Combined), HPSv2标签定义
             \rowcolor{gray!10} \cellcolor{white} \multirow{-5}{*}{\textbf{HPSv2}} & \textbf{Pro-GRPO (HPS+CLIP)} & \multicolumn{2}{c}{\textbf{0.596}} & \textbf{6.273} & \textbf{1.140} & \textbf{22.271} & \textbf{0.3800} \\
            \bottomrule
        \end{tabular}
    }
\end{table*}

\noindent\textbf{Performance on flow-based models.}
Table~\ref{tab:flow_results} demonstrates that Pro-GRPO improves both optimization objectives and out-of-domain (OOD) metrics while reducing training costs.
On DrawBench, Pro-GRPO consistently outperforms the baseline Flow-GRPO across all metrics with a PickScore of 24.008 (+0.686) and Aesthetic score of 6.046 (+0.134). Notably, as shown in the `Efficiency' column, this performance gain is achieved alongside a 1.26$\times$ training speedup, attributed to the reduced group size during optimization.
This advantage extends to HPSv2, as reflected in OOD metrics like ImageReward (1.509 vs. 1.463).
The efficiency benefits are particularly evident in the lightweight Pro-GRPO-Flash ($G_{\text{max}}=24 \to 12$). It delivers a 1.41$\times$ speedup while still surpassing the baseline on DrawBench (23.722 vs. 23.322). This result isolates the benefit of in-process pruning, supporting our ``less-is-more'' hypothesis.
Complementing these metrics, Fig.~\ref{fig:training_curves} (a) shows faster convergence and a higher asymptote. Together with the qualitative comparisons in Fig.~\ref{fig:qualitative_results}, these findings indicate that Pro-GRPO improves sample efficiency and robustness without overfitting.

\noindent\textbf{Performance on diffusion-based models.}
Table~\ref{tab:diffusion_results} and Fig.~\ref{fig:training_curves} (b,c) present the comparison against DanceGRPO.
In the single-objective setting (HPSv2.1), Pro-GRPO demonstrates enhanced optimization capability, surpassing DanceGRPO in in-domain rewards (0.391 vs.\ 0.369) and consistently improving OOD metrics (e.g., ImageReward on HPSv2: 1.114 vs.\ 1.079). This is mirrored in Fig.~\ref{fig:training_curves}(b) by a faster rise and higher asymptote.
In the multi-objective setting (HPSv2.1\,+\,CLIP), the advantage of our approach is also pronounced. Pro-GRPO achieves the best OOD results across the board on both datasets. Fig.~\ref{fig:training_curves} (c) shows a stable margin throughout training on the composite objective.
These findings support our core insight: a small, high-variance survivor set provides a stronger learning signal than the full, unfiltered group.

\noindent\textbf{Semantic alignment and compositionality.}
We assess text–image grounding on GenEval (Table~\ref{tab:geneval_results}), which is not used for training (all models are tuned only with PickScore). Pro-GRPO surpasses Flow-GRPO on the Overall score (0.726 vs.\ 0.719) and on every fine-grained category: Two-Obj.\ (0.947 vs.\ 0.942), Colors (0.867 vs.\ 0.851),  Position (0.323 vs.\ 0.288, +0.035). The strongest gain on Position indicates improved spatial reasoning and compositional binding, suggesting that variance-aware pruning produces trajectories that carry clearer semantic signals. The consistent improvements on this unseen benchmark further support that Pro-GRPO enhances semantic alignment without reward-specific overfitting.

\noindent\textbf{Computational efficiency analysis.}
\label{sec:efficiency}
To quantify the cost-effectiveness of Pro-GRPO, we use calflops~\cite{calflops} to profile per-call FLOPs and the cost of a full sampling–training cycle (Table~\ref{tab:efficiency_flops}).
Pro-GRPO reallocates compute from uniform training on a fixed group to exploration and selection. It briefly expands the initial pool ($G_{\max}$) and prunes in-process, early-terminating reward-clustered trajectories to avoid their remaining denoising steps. Because only the survivor set $K$ proceeds to backprop, the optimization cost scales with $K$ rather than $G_{\max}$.
This yields substantial savings. FLOPs drop from 453474\,T (Flow-GRPO) to 335627\,T with Pro-GRPO Standard (–26\%, 1.26$\times$ speedup), and to 267366\,T with Pro-GRPO Flash (–41\%, 1.41$\times$). The performance in Table \ref{tab:flow_results} shows that investing compute in exploration and variance-based pruning is more effective and more efficient than uniformly training on a large, unfilter group. The detailed calculation procedure is provided in the appendix.

\begin{table}[t]
    \centering
    \caption{\textbf{Quantitative evaluation on GenEval benchmark.} All models are fine-tuned using the PickScore reward. We report the Overall score and fine-grained accuracies.}
    \label{tab:geneval_results}
    \resizebox{\columnwidth}{!}{
        \begin{tabular}{lcccc}
            \toprule
            \multirow{2}{*}{\textbf{Model}} & \textbf{Overall} & \multicolumn{3}{c}{\textbf{Fine-grained Accuracy}} \\
            \cmidrule(lr){3-5}
             & Score ($\uparrow$) & Two Obj. & Colors & Position \\
            \midrule
            SD 3.5-M (Base) & 0.696 & 0.869 & 0.817 & 0.248 \\
            Flow-GRPO & 0.719 & 0.942 & 0.851 & 0.288 \\
            % Standard 版本：Colors 和 Position 最优
            \rowcolor{gray!10} \textbf{Pro-GRPO} & \textbf{0.726} & \textbf{0.947} & \textbf{0.867} & \textbf{0.323} \\
            \bottomrule
        \end{tabular}
    }
\end{table}

\begin{table}[t]
    \centering
    \caption{\textbf{Computational efficiency analysis.} We report the FLOPs (in Tera) and relative speedup. The top section lists the cost of atomic operations. The bottom section compares the total computational cost of a full epoch, including trajectory sampling and policy optimization. Pro-GRPO reduces the aggregate FLOPs by early-terminating reward-clustered trajectories during the sampling phase.}
    \label{tab:efficiency_flops}
    \resizebox{\columnwidth}{!}{
        \begin{tabular}{lcc}
            \toprule
            \textbf{Component / Method} & \textbf{FLOPs (T) $\downarrow$} & \textbf{Speedup $\uparrow$} \\
            \midrule
            \multicolumn{3}{l}{\textit{Atomic Operations (Per Call)}} \\
            \quad Noise Prediction & 3.88 & -- \\
            \quad VAE Decoding & 2.49 & -- \\
            \quad Reward Computation & 0.34 & -- \\
            \midrule
            \multicolumn{3}{l}{\textit{Full Training Framework}} \\
            \quad Flow-GRPO (Baseline) & 453474.18 & 1.00$\times$ \\
            % Pro-GRPO Standard - Gray background
            \rowcolor{gray!10} \quad \textbf{Pro-GRPO (Standard)} & 335626.82 & \textbf{1.26$\times$} \\
            % Pro-GRPO Flash - Gray background
            \rowcolor{gray!10} \quad \textbf{Pro-GRPO (Flash)} & \textbf{267365.79} & \textbf{1.41$\times$} \\
            \bottomrule
        \end{tabular}
    }
\end{table}

\subsection{Ablation Studies}
\label{sec:ablation}

\noindent\textbf{Effect of Initial Group Size.}
Table~\ref{tab:ablation_scaling} investigates the impact of the initial group size $G_{\text{max}}$ on performance. By fixing the final survivor set $K=8$ and scaling $G_{\text{max}}$ from 32 to 64, we observe a consistent improvement across all metrics by exposing superior candidates for OVF. 
We also observe a trend of diminishing rewards: the performance gains taper off at larger scales (e.g., $48 \to 64$). The Out-of-Domain HPSv2.1 score plateaus at 0.383, indicating that the marginal informational value of additional trajectories saturates. This suggests that scaling $G_{\max}$ indefinitely yields limited benefits beyond a sufficiency threshold.
Because Pro-GRPO prunes in-process, the effective optimization cost remains tied to $K$ rather than $G_{\text{max}}$, enabling us to exploit larger initial group size without proportional compute growth.

\noindent\textbf{Effect of Pruning Timesteps.}
Table~\ref{tab:ablation_timesteps} examines where to place pruning checkpoints under \(T{=}50\) with a path \(32\!\rightarrow\!8\).
The results show that OVF-based pruning is effective across schedules; even an early schedule \(\{10,20\}\) yields gains.
Shifting the checkpoints deeper improves accuracy, with \(\{30,40\}\) attaining the best trade-off. For example, in-domain HPSv2.1 rises from 0.373 to 0.391, and OOD PickScore from 21.711 to 22.075.
The pattern suggests that an early checkpoint can remove clear low-value samples for compute savings, while a later checkpoint refines the survivor set once latents carry richer semantics.

\begin{table}[t]
    \centering
    \caption{\textbf{Ablation study on scaling initial group size ($G_{\text{max}}$).} Experiments are conducted on SD1.4 optimized with HPSv2.1.}
    \label{tab:ablation_scaling}
    \resizebox{\columnwidth}{!}{
        % 1. 将原来的 lccccc 改为 cccccc，让第一列居中
        \begin{tabular}{cccccc} 
            \toprule
            \multirow{2}{*}{\textbf{Pruning Path}} & \textbf{In-Domain} & \multicolumn{4}{c}{\textbf{Out-of-Domain}} \\
            \cmidrule(lr){2-2} \cmidrule(lr){3-6}
            % 第一列居中后，这里的 ($G_{\max} \to K$) 也会自然居中
             ($G_{\text{max}} \to K$) & HPSv2.1 & Aes. & IR & Pick. & HPSv2.1 \\
            \midrule
            $32 \to 8$ & 0.386 & 6.327 & 1.080 & 22.075 & 0.379 \\
            % 2. 删除了 (Standard)
            $48 \to 8$ & 0.391 & 6.253 & 1.114 & 22.202 & \textbf{0.383} \\
            % 灰色高亮表现最好的一行
            \rowcolor{gray!10} $64 \to 8$ & \textbf{0.393} & \textbf{6.350} & \textbf{1.120} & \textbf{22.231} & \textbf{0.383} \\
            \bottomrule
        \end{tabular}
    }
\end{table}

\begin{table}[t]
    \centering
    \caption{\textbf{Ablation study on pruning checkpoints.} Experiments are conducted on SD-v1.4 (\(T=50\)) optimized with HPSv2.1, using a pruning path of \(32 \to 8\).}
    \label{tab:ablation_timesteps}
    \resizebox{\columnwidth}{!}{
        \begin{tabular}{cccccc}
            \toprule
            \multirow{2}{*}{\textbf{Pruning Steps}} & \textbf{In-Domain} & \multicolumn{4}{c}{\textbf{Out-of-Domain}} \\
            \cmidrule(lr){2-2} \cmidrule(lr){3-6}
             ($t_{1}, t_{2}$) & HPSv2.1 & Aes. & IR & Pick. & HPSv2.1 \\
            \midrule
            $10, 20$ & 0.373 & 6.132 & 1.016 & 21.711 & 0.364 \\
            $20, 30$ & 0.382 & 6.160 & 1.063 & 21.859 & 0.369 \\
            % 灰色高亮表现最好的一行 (Ours)
            \rowcolor{gray!10} $30, 40$ (Ours) & \textbf{0.391} & \textbf{6.327} & \textbf{1.080} & \textbf{22.075} & \textbf{0.379} \\
            \bottomrule
        \end{tabular}
    }
\end{table}

\section{Conclusion}
\label{sec:conclusion}
In this work, we address the computational bottleneck in GRPO caused by the conflict between large group sizes and cost. Identifying a "Reward Clustering Phenomenon" that dilutes advantage signals, we initially explored Optimal Variance Filtering (OVF). While OVF confirms that small, high-variance subsets are superior, its post-sampling nature incurs overhead. Consequently, we propose Pro-GRPO, a framework shifting to proactive, in-process pruning. By leveraging latent features to early-terminate redundant samples, Pro-GRPO’s ``Expand-and-Prune'' strategy decouples exploration breadth from optimization cost. Experiments on diffusion and flow-based models demonstrate significantly enhanced alignment performance and efficiency.

{
    \small
    \bibliographystyle{ieeenat_fullname}
    \bibliography{main} % <-- 参考文献现在在最后
}

\appendix
% 1. 在参考文献之前加载附录
% 2. 建议在附录和参考文献之间加一个 \clearpage 
%    来确保参考文献从新的一页开始，这在审阅时更整洁


\end{document}